\documentclass{article}

\usepackage{arxiv}

\usepackage[utf8]{inputenc}
\usepackage[T1]{fontenc}
\usepackage{hyperref}
\hypersetup{
colorlinks=true,
linkcolor=blue,
citecolor=blue,
urlcolor=blue
}
\usepackage{url}
\usepackage{booktabs}
\usepackage{amsmath}
\usepackage{amsfonts}
\usepackage{microtype}
\usepackage{graphicx}
\usepackage{natbib}

\usepackage{xcolor}
\usepackage[table]{xcolor}

\usepackage{booktabs}
\usepackage{multirow}

\title{Automated Design Optimization via Strategic Search with Large Language Models}

\author{
  Anthony Carreon\thanks{Corresponding author. Email: \texttt{acarreon@umich.edu}}~~, Vansh Sharma, and Venkat Raman \\
  Department of Aerospace Engineering\\
  University of Michigan\\
  Ann Arbor, MI 48109 \\
}

\begin{document}
\maketitle

\begin{abstract}
Optimization methods have long advanced many fields, yet they struggle when faced with design problems where the search space and design parameters are difficult to define. Large language models (LLMs) offer a promising alternative by dynamically interpreting design spaces and leveraging encoded domain knowledge. To this end, we present AUTO: an iterative optimization framework that treats design optimization as a strategic search guided by LLM reasoning. The framework separates high-level planning by a Strategist agent from low-level implementation by concurrent Implementor agents, iteratively refining designs through explore-exploit strategies. We demonstrate AUTO on three GPU code optimization problems. For chemical kinetics, AUTO outperforms in-lab-optimized code by up to 1.74$\times$ for problem sizes up to $10^5$ cells. For matrix multiplication, AUTO achieves up to 94\% of cuBLAS double-precision performance. For KernelBench, we achieve speedups of up to 118$\times$ over PyTorch baselines across 29 problems spanning individual operators and full neural network architectures; however, cheating was frequently observed. A posteriori analysis reveals 50~--~70\% alignment with Bayesian optimization sampling strategies. All AUTO simulations ran within 100 iterations (about 10 hours), with estimated costs of \$15~--~159 per run for proprietary models. Furthermore, AUTO is built entirely on open-source LLMs and libraries, demonstrating affordability and data privacy. Given AUTO's generizability and flexibility, future work will explore domains beyond GPUs and supercomputing.
\end{abstract}


\section{Introduction}

Traditional optimization has shown strong results in various fields, from gradient descent in machine learning to Bayesian optimization for materials discovery \citep{zhang2020bayesian} and chemical process design \citep{li2021nextorch}, to evolutionary algorithms for antenna design \citep{hornby2006automated}. However, these methods struggle in vast, ill-defined search spaces with complex constraints. Gradient-free methods, including evolutionary algorithms and Bayesian optimization, work when the design space is parameterizable but are incompatible with problems where defining transformations is difficult or parametrizations are unclear. 

Recent advances in large language models (LLMs) offer a promising alternative by dynamically interpreting design spaces and leveraging domain knowledge encoded in their training data or through semantic search. These advances have led to systems like FunSearch \citep{romeraparedes2023funsearch}, which discovered new mathematical solutions through evolutionary LLM-driven search, and AlphaEvolve \citep{novikov2025alphaevolve}, which discovered new solutions to diverse computational problems. Building on these advances, we introduce AUTO, an LLM agent framework that treats design optimization as a strategic search problem. Complementary to hypothesis-driven approaches focused on high-level system design \citep{hamadanian2025glia}, AUTO targets implementation-level optimization by exploring the design space. Our framework draws inspiration from evolutionary algorithms and Bayesian optimization by iteratively refining designs through (a) improving existing candidates, (b) combining successful patterns, (c) innovating when existing approaches plateau, or (d) reattempting failed designs with targeted corrections.

We demonstrate AUTO on GPU code optimization, a domain relevant to fields from machine learning to scientific computing. Recent work has explored LLM-based code optimization through various approaches. \citet{afterburner2025} demonstrated reinforcement learning methods for competitive programming, while \citet{gpukernelscientist2025} targeted scenarios with limited hardware documentation, and \citet{fortrankokkos2025} automated legacy code transformation using multi-agent systems. AUTO extends these efforts by framing optimization as a systematic search problem, thereby aligning with established optimization methods. By demonstrating AUTO's ability to autonomously discover competitive GPU implementations, this work opens avenues towards democratizing HPC engineering. Beyond HPC, there are optimization applications (not shown here) spanning scientific simulation, research workflows, and multidisciplinary systems optimization. Furthermore, the framework is built entirely on open-source LLMs and libraries, demonstrating affordability and data privacy.

The remainder of this paper is organized as follows: Section~\ref{sec_opt_framework} describes the AUTO framework architecture and optimization workflow. Section~\ref{optimizer_apps} presents the GPU programming applications and implementation details. We analyze solution quality, search efficiency, and cost-effectiveness in Section~\ref{sec_results_discussion}, followed by conclusions and future directions in Section~\ref{sec_conclusion}.

\section{The LLM-Based Optimizer Framework (AUTO)}\label{sec_opt_framework}

\begin{figure}[h]
    \centering
    \includegraphics[width=\linewidth]{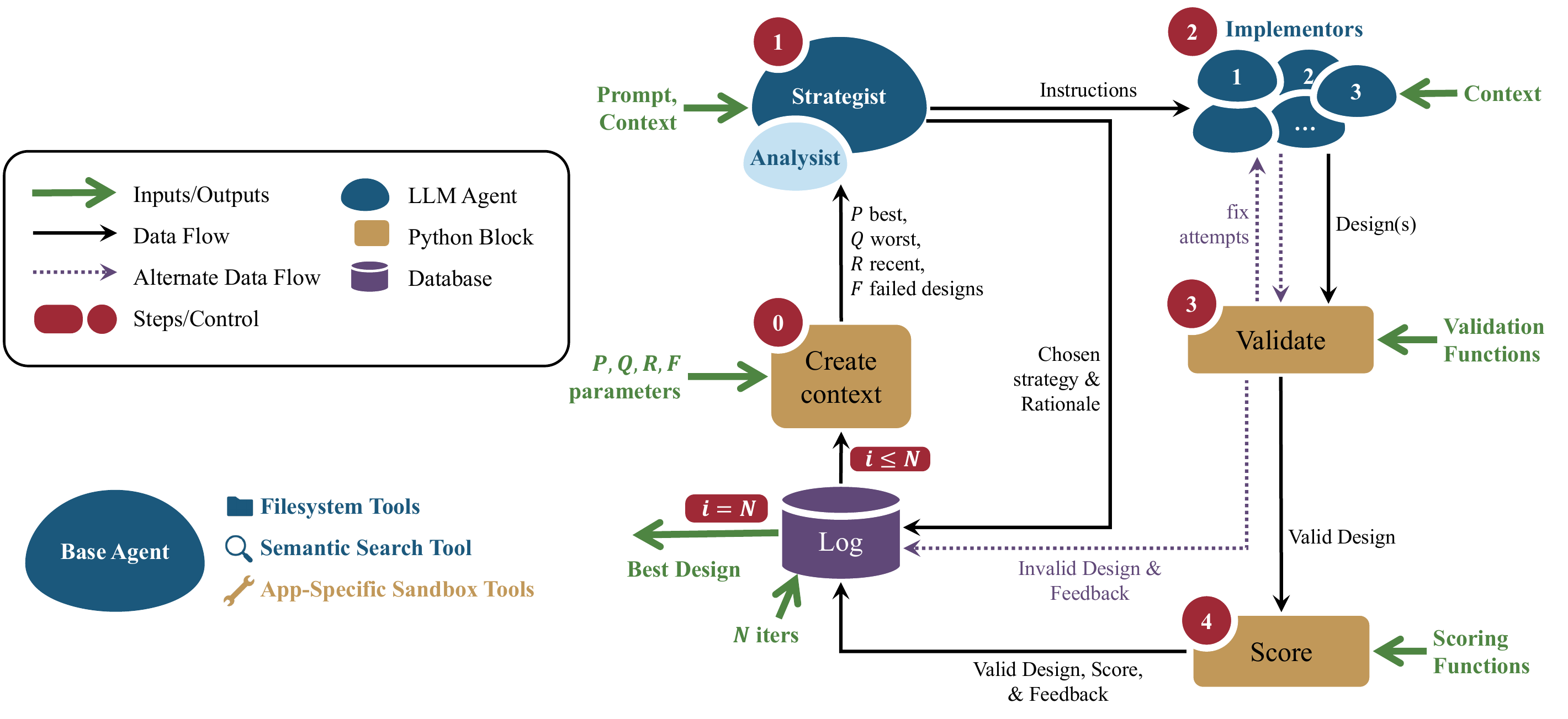}
    \caption{AUTO: An LLM-based design optimization framework. The optimization loop runs for up to $N$ iterations. (0) Create context from historical design data to inform subsequent iterations. (1) The Strategist, supported by an Analyst, selects an optimization strategy. (2) Multiple Implementors generate designs from strategic instructions. (3) Validate designs and return to Implementors for fixes if needed; (4) Score valid designs. The cycle repeats for up to $N$ iterations, terminating early if no improvement in score is observed after a configurable number of iterations.}
    \label{fig_framework}
\end{figure}

AUTO employs LLMs as agents in a collaborative environment. Inspired by gradient-free optimization techniques such as Bayesian optimization and evolutionary algorithms, the framework iteratively improves designs by learning from a population of previously generated candidates. All agents in AUTO inherit from a common base agent, adopted from OS-level memory management paradigms \citep{packer2023memgpt}, that provides a sandboxed work directory with a filesystem toolset, automatic chat history summarization, semantic and keyword search of documentation (if provided), and planning tools for focused work. Some agents can further spawn child agents, which are themselves instances of this same base agent. High-level planning by the Strategist agent is separate from low-level implementation by the Implementor agents. This separation breaks the optimization problem into manageable tasks while circumventing LLM context-window limitations. The Strategist may call on an Analyst child agent that examines existing implementations and documentation before the Strategist makes its decision. Multiple Implementors operate concurrently within each iteration, each independently generating, validating, and scoring designs as instructed by the Strategist. Figure~\ref{fig_framework} shows the complete workflow, and subsequent paragraphs describe each step in detail.

\paragraph{Step 0: Context Curation from Logged Data}

Each design iteration is added to a growing database, including its strategies, validation status, detailed evaluation metrics, and agent interactions. This logging enables analysis of the optimization trajectory and is analogous to surrogate-model updating in Bayesian optimization. The design space is represented by the set of all designs up to the start of iteration $t$, $\mathcal{D}_t$ (i.e., all designs before new ones generated during iteration $t$). This set includes both successful and unsuccessful designs (those that failed validation or scoring or are considered "cheating"). During context curation at step 0 in Figure~\ref{fig_framework}, the goal is to construct a subset of designs, $\mathcal{C}_t \subseteq \mathcal{D}_t$, that when presented to the Strategist, leads to an informed decision. This construction must balance presentation detail and richness with constraints on the context window size and LLM accuracy. We reason that, by composing $\mathcal{C}_t$ with a small mixture of $P$ best, $Q$ worst, $R$ recent, and $F$ randomly selected failed designs, the Strategist's decision will yield a design inspired by the top-scoring ones, improved from the bottom-scoring ones, diversified from the recent ones, and/or corrected from the failed ones. In highly nonlinear, non-smooth design surfaces, bottom-scoring designs may be adjusted or learned from to achieve significant improvements; hence, tagging along with $Q$ of the worst-scoring designs may be beneficial. Similarly, including $F$ failed designs provides the Strategist with information about constraint violations and dead-end approaches, enabling the Strategist to avoid or reattempt them with targeted fixes. The sizes of the sets $P$, $Q$, $R$, and $F$ are configurable and disjoint to avoid redundancy in the context window.

\paragraph{Step 1: The Strategist}\label{par_strategist_agent}
The Strategist is tasked with analyzing the problem context, sketches, design hints, documentation, and $\mathcal{C}_t$. The problem context is background information about the problem being optimized. The sketches may be existing designs or partial designs related to the application that serve as starting design points. Design hints may include possible actions, design parameters, or guidelines for manipulating and exploring the design space. Before making its decision, the Strategist delegates to an Analyst child agent that reads and compares existing implementations and explores the documentation for relevant information. Based on the Analyst's findings and the curated context, the Strategist must select one of four predefined sampling strategies, providing evidence and rationale, and provide detailed implementation instructions to the Implementors. The four sampling strategies are:

\begin{enumerate}
    \item "\textbf{innovate}" is chosen when all or many implementations exhibit similar designs and scores, when a distinctly different approach could yield breakthrough improvements, or when recent implementations fail to satisfy constraints with no obvious fixes.
    \item "\textbf{combine}" is selected when multiple implementations excel in different aspects, when some implementations have strong design patterns in one dimension while others excel in a different dimension, when hybrid approaches can capture benefits from multiple designs, or when different design approaches are complementary.
    \item "\textbf{refine}" is chosen when an implementation is close to reaching a breakthrough score, when specific bottlenecks are identified and fixable, when constraint-related errors are minor and correctable, or when the core design is sound but needs tuning or fixing.
    \item "\textbf{re-attempt failure}" is chosen when a previous implementation failed validation but shows clear potential for fixes and improvements, when the failure mode is well-understood and correctable, or when the underlying design approach is promising despite constraint violations.
\end{enumerate}

The system prompt also guides the Strategist in presenting information to the Implementors. For instance, the Strategist must provide specific instructions (e.g., "\textit{use 32$\times$32 tiles in shared memory with column-major layout}" instead of "\textit{implement optimal memory access}"). It should also consider optimization hints and avoid conflicting with the user's instructions and requirements.

\paragraph{Steps 2, 3, and 4: The Implementors, Validation, and Scoring}

Multiple Implementors operate concurrently within each iteration, each independently generating designs from the Strategist's instructions and fixing constraint errors as they occur. These constraints are domain-specific. For code optimization, this includes compilation, execution, and testing against ground-truth data points. For other design tasks, validation may involve meeting manufacturing constraints or regulatory requirements. As with any optimization algorithm, context on all possible design parameters and their values is needed to find an implementation that meets the design constraints; thus, the Implementors are provided with domain-specific documentation accessible via semantic search. If any check fails, the Implementor automatically receives feedback and retries within its persistent context window. If the Implementor's token usage exceeds a configurable limit, AUTO skips step 4 (scoring) and logs the Implementor's latest design and its metadata, along with the failure reason. The Strategist may leverage failed designs during Step 1 if they are among the $F$ failures or the $R$ most recent designs.

Valid designs are evaluated against problem-specific objectives. The score consists of a scalar value, its units, objective (maximize or minimize), and an optional detailed breakdown of the scoring. For instance, the GPU optimization problems presented in this work use the total GPU kernel runtime as the score, measured in milliseconds, and aim to minimize it. Its detailed breakdown consists of the runtime for each kernel and reports from NVIDIA Nsight Systems. Ideally, the scoring process measures relevant performance metrics across multiple operating conditions. The scalar score is used during context curation in Step 0; however, the Strategist is presented with full evaluation details for each design in the curated context to support informed decision-making.

\section{Optimizer Applications}\label{optimizer_apps}

To evaluate AUTO's capabilities, we apply the framework to GPU code optimization. High-performance computing increasingly relies on GPU acceleration, from deep learning \citep{sharma2025accelerating} to fluid dynamics simulation \citep{sharma2024amrex, carreon2025gpu}. However, optimizing GPU kernels demands specialized expertise in hardware architecture, memory hierarchies, and parallel programming. To this end, we evaluate AUTO on two tasks with differing performance characteristics: chemical kinetics integration and dense matrix multiplication. These applications test AUTO's ability to discover optimizations competitive with expert-tuned implementations.

\subsection{Chemical Kinetics}

We deploy AUTO to optimize the performance of a chemical kinetics GPU code. Specifically, AUTO must optimize a CUDA program that integrates millions of chemical reaction systems in parallel. This problem is relevant to compressible combustion in computational fluid dynamics (CFD) and is a primary computational bottleneck in many combustion applications \citep{raman2023nonidealities}. We consider the Robertson problem \citep{robertson1966}, a classic three-species, three-reaction chemical system:

\begin{align}
\text{A} &\xrightarrow{0.04} \text{B} \\
\text{B} + \text{B} &\xrightarrow{3 \times 10^7} \text{C} + \text{B} \\
\text{B} + \text{C} &\xrightarrow{1 \times 10^4} \text{A} + \text{C}
\end{align}

This yields the following system of ordinary differential equations:

\begin{align}
\dot{x} &= -0.04x + 10^4 yz \\
\dot{y} &= 0.04x - 10^4yz - 3 \times 10^7y^2 \\
\dot{z} &= 3 \times 10^7y^2
\end{align}

The ODE system exhibits timescales spanning several orders of magnitude, leading to highly variable iteration counts during integration across different initial conditions. The LLM's task is to generate an optimal GPU implementation to time-integrate $N$ such systems with different initial conditions, mirroring the variation in thermochemical states across millions of computational cells in combustion CFD\footnote{A note on terminology: we refer to a single ODE system as a "cell" for brevity and relevance to combustion CFD applications.}. Conditional on the implementation approach, GPU performance bottlenecks may include, but are not limited to, warp divergence, memory coalescence, and kernel-launch overhead.

\subsection{Dense Square Matrix Multiplication}

We also deploy AUTO to optimize dense matrix multiplication performance on GPUs. This problem serves as a baseline benchmark and is ubiquitous across scientific computing, machine learning, and numerical methods. We consider the dense, $N\times N$ matrix multiplication problem, $C = A \times B$, where all matrices are randomly generated with values between 0 and 1 and stored in column-major format in double precision. AUTO's task is to search for an optimal CUDA implementation across a range of problem sizes, from $N=128$ to $N=8{,}192$, without access to vendor-optimized libraries such as cuBLAS or cuSparse. The implementation is restricted to raw CUDA, PTX, or CUTLASS. The inherent data reuse in this operation makes it particularly amenable to GPU optimization through techniques such as tiling, memory hierarchy exploitation, and thread-level parallelism \citep{volkov2008benchmarking}. We compare AUTO's solutions against the double-precision general matrix multiplication subroutine (GEMM), \texttt{dgemm()}, in NVIDIA's cuBLAS library \citep{cublas}, which represents years of optimization efforts by teams of human experts and serves as the "ground truth" for dense matrix multiplication on NVIDIA hardware.

\subsection{KernelBench GPU Kernel Optimization}

We also evaluate AUTO on KernelBench \citep{kernelbench2025}, a benchmark for assessing LLMs' ability to generate efficient GPU kernels from PyTorch reference implementations. KernelBench is organized into four levels of increasing complexity: Level 1 (100 problems) targets individual operators such as matrix multiplication variants, convolutions, and activation functions; Level 2 (100 problems) targets simple operator fusion patterns; Level 3 (50 problems) encompasses complete neural network architectures such as AlexNet, ResNet, and Mamba; and Level 4 (20 HuggingFace model architectures), which we do not evaluate here. For each problem, the LLM is provided with a reference PyTorch \texttt{nn.Module} and must produce a \texttt{ModelNew} class that implements the same computation using custom CUDA kernels compiled at runtime via \texttt{torch.utils.cpp\_extension.load\_inline}. Performance is measured as speedup over the PyTorch baseline. We evaluate AUTO on 53 randomly selected problems across levels 1~--~3.

\section{Implementation Details}\label{sec_implementation_details}

The framework is written in Python using the Pydantic AI framework~\citep{pydantic_ai} with Ollama~\citep{ollama} for multiple LLM servers. Each GPU is mapped to a single Ollama server running a single LLM. GPT-OSS-120b~\citep{agarwal2025gpt} is used as the base LLM for the Strategist, Analyst, and Implementors. In this work, a single AUTO run uses multiple H100 GPUs from a local lab cluster, which are dynamically partitioned into two pools: one pool hosts the Ollama servers, while the other hosts code-execution sandboxes (one sandbox per GPU) for compiling, correctness checking, and performance profiling. This partitioning maintains exclusivity and avoids interference between LLM inference and code execution. The AUTO framework runs on the CPU, manages both GPU pools, and handles I/O. In each of the applications presented here, two GPUs are used for the sandboxes, while 3 or 4 GPUs are used for the Ollama servers. The sandboxes are checked out with async-safe locks to agents for tool calls or to the validation and scoring pipeline, as needed, and are immediately released. This ensures efficient and fair usage of the limited sandboxes, avoids contention, and prevents race conditions.

To enable longer agent runs, chat histories are automatically compressed using a minimal \textit{Summarizer} agent when the LLM's context window is approaching its limit. Additionally, tool outputs exceeding a configurable length threshold (e.g., verbose compiler logs) are summarized by the Summarizer agent before being returned to the agent that issued the tool call, preventing potential derailment of the agent's focus and reasoning.

The knowledge base backend is a custom persistent vector index created from multiple user-specified filesystem paths, with structure-aware chunking, hashing, and embedding using the Sentence Transformers library \citep{reimers2019sentencebert}. In this work, we use the Jina embeddings 2 model \citep{gunther2023jina}. Agents can query the knowledge base (based on cosine similarity) and use command line tools (such as \texttt{ls} and \texttt{grep} with regular expressions) to further inspect the filesystem contents. In earlier iterations of this work, this knowledge base was unavailable, resulting in a significant number of unrecoverable compilation failures. As a remedy, the current work equips the agents with a curated knowledge base of CUDA, PTX, and CUTLASS documentation.

\begin{figure}
    \centering
    \includegraphics[width=\linewidth]{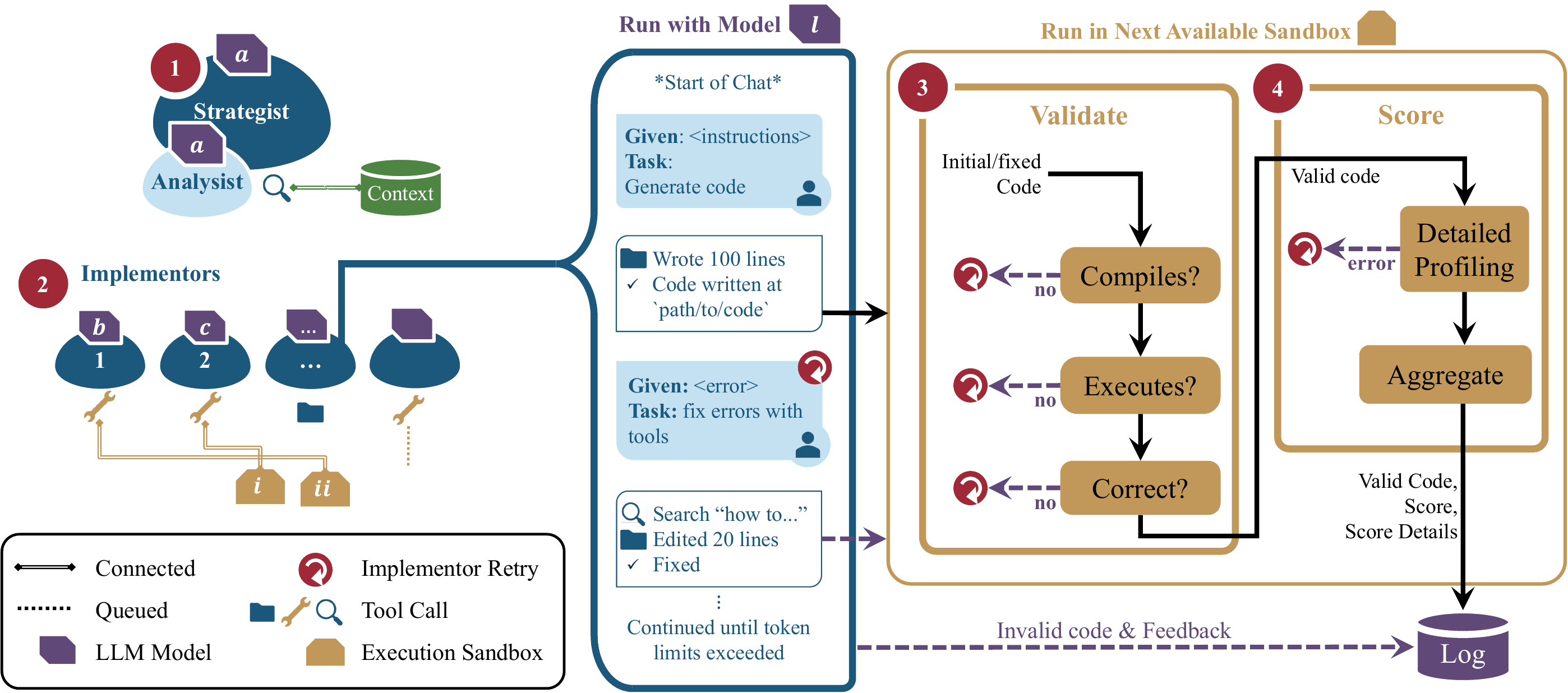}
    \caption{Implementation details of Steps 1~--~4 from Figure~\ref{fig_framework} as it applies to GPU code optimization. The "Validate" and "Score" steps are Python blocks of code executed in sandboxed SLURM GPU jobs, while interactions with the agents are handled by LLMs, each mapped to a GPU. Sandboxes are checked out with async-safe file locking to avoid race conditions for every agent tool call that needs execution permission.}
    \label{fig_imp_details}
\end{figure}

Table~\ref{tab_exp_config} summarizes the parameters for all applications, while
Figure~\ref{fig_imp_details} visualizes Steps 1~--~4 from Figure~\ref{fig_framework} as it applies to GPU code optimization. In step 2, the Implementors are required to generate code with a specific input and output interface to ensure compatibility with the validation and scoring steps. In step 3, the generated code is checked against three validator functions: compilation, execution, and correctness. Validators are configured by the user as a dependency graph to determine their execution order in the validation pipeline. For example, the correctness check depends on the execution step, which in turn depends on the compilation step. To improve LLM token efficiency, all possible validators are executed, skipping failures while preserving the dependency order. Any failures are collected and returned in one LLM call to the appropriate Implementor for fixes within its persistent context window. If the Implementor exhausts its token limit, the implementation is recorded as failed, and the scoring step is skipped.

\begin{table}[h]
\centering
\caption{AUTO parameters for all applications.}
\renewcommand{\arraystretch}{1.5}
\small
\begin{tabular}{>{\raggedleft\arraybackslash}p{0.17\linewidth}>{\centering\arraybackslash}p{0.25\linewidth}>{\centering\arraybackslash}p{0.25\linewidth}>{\centering\arraybackslash}p{0.25\linewidth}}
\toprule
\textbf{Parameter} & \textbf{Chemical Kinetics} & \textbf{Matrix Multiplication} & \textbf{KernelBench} \\
\midrule
Base LLM & \multicolumn{3}{c}{GPT-OSS-120b} \\
\hline
Implementors per iteration. & 4 & 3 & 3 \\
\hline
Max Iterations & \multicolumn{3}{c}{100} \\
\hline
Strategist Context (Section~\ref{sec_opt_framework}) & \multicolumn{3}{c}{$P\!=\!5$, $Q\!=\!3$, $R\!=\!3$, $F\!=\!2$} \\
\hline
Correctness Check & 200 cells at $t_\text{end}\!=\!0.5$ with up to $5\!\times\! 10^5$ iter.\ per cell. & $N\!\times\! N$ matrices with $x \sim U(0,1)$ for $N\!=\!32$, $128$, $512$. & Reference PyTorch output for benchmark-defined inputs. \\
\hline
Performance Evaluation & Total kernel runtime for $10^5$ cells at $t_\text{end}\!=\!10$ with up to $10^6$ iter.\ per cell. & Total kernel runtime for $N\!=\!4{,}096$. & Speedup over PyTorch baseline. \\
\hline
Knowledge Base & \multicolumn{3}{c}{CUDA, PTX, CUTLASS documentation} \\
\bottomrule
\end{tabular}
\label{tab_exp_config}
\end{table}

The kinetics and matrix multiplication applications use reference data for validation across multiple operating conditions (see Table~\ref{tab_exp_config}), while KernelBench uses the original benchmark's built-in validation. In step 4 (scoring), valid implementations are scored according to the application-specific objectives presented in Table~\ref{tab_exp_config}. Due to the long run time required by NVIDIA Nsight Systems for detailed performance timing, only a representative problem size is used for scoring. The total kernel runtime is recorded as the design score with the objective of minimizing it. In KernelBench, the score is defined as the speedup relative to the PyTorch baseline, and the objective is to maximize it. Solutions with speedups exceeding 10$\times$ (designs too good to be true) trigger the Analyst agent, which inspects the code for cheating (e.g., delegating to the reference PyTorch model or bypassing the GPU benchmark with a no-op kernel). This was an observed phenomenon in some KernelBench runs and iterations and reported by \citet{kernelbench2025}.

\section{Results and Discussion}\label{sec_results_discussion}

The discussion and analysis of the results are split into three subsections. The first subsection analyzes AUTO's solutions for matrix multiplication, kinetics, and KernelBench. Next, we explore AUTO's performance as an optimization framework by drawing comparisons to Bayesian optimization. The final subsection discusses the costs and benefits of the optimization framework.

\newpage

\subsection{Solution Quality}

\subsubsection{Kinetics and Matrix Multiplication}

\begin{figure}[h]
    \centering
    \includegraphics[width=\textwidth]{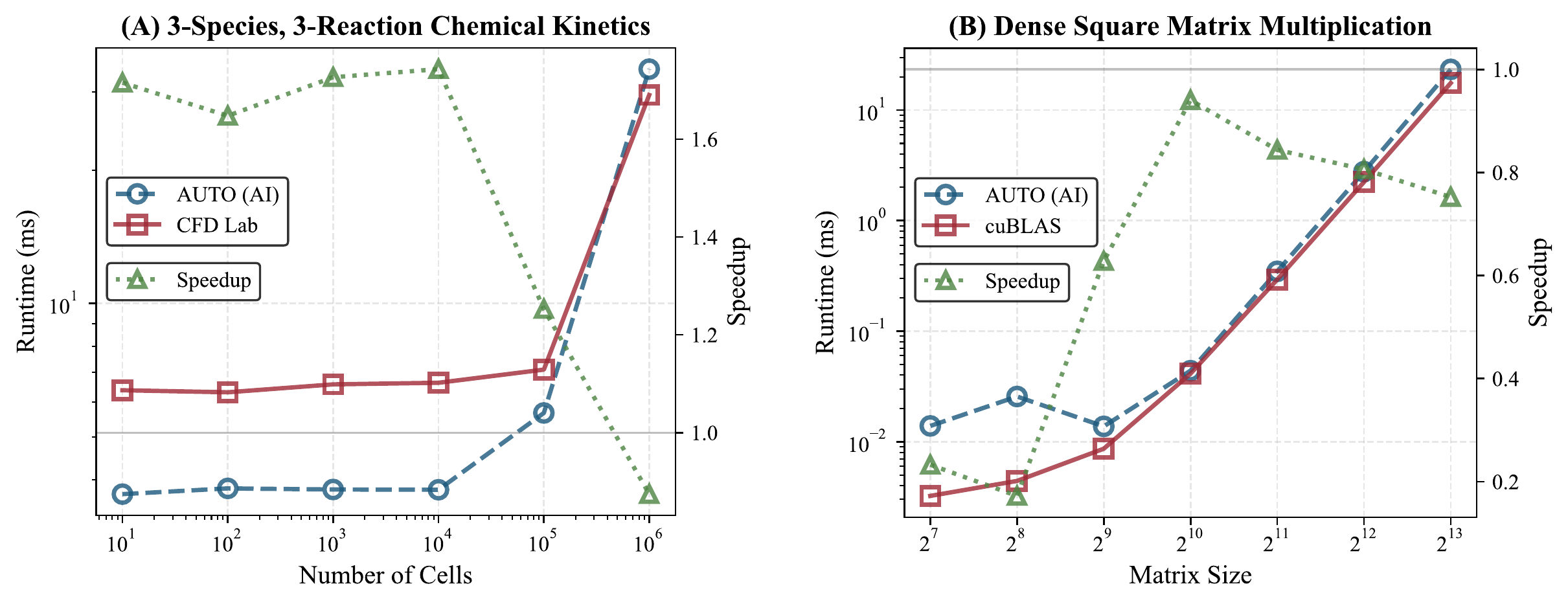}
    \caption{Timing comparisons between the agent-optimized and human-optimized GPU code for (A) the kinetics problem by \citet{robertson1966} across different problem sizes and (B) matrix multiplication applied to two $N\times N$ matrices. See Table~\ref{tab_exp_config} for the parameters of each application.}
    \label{fig_benchmark_all_problems}
  \end{figure}

Figure \ref{fig_benchmark_all_problems} compares in-house optimized GPU code against AUTO's best solutions for (A) Robertson chemical kinetics and (B) square matrix multiplication. AUTO's kinetics solution outperforms the in-house optimized CFD lab code by $\sim 1.6 \times$ for problem sizes 10~--~$10^5$ cells. At $10^6$ cells, the AUTO solution is slightly slower ($\sim 0.9 \times$). The AUTO solution wins at small cell counts, likely due to lower fixed overhead costs (e.g., compact data layout, pinned memory, async streams), whereas the manually-optimized code wins at 1M cells, likely due to its early-exit logic in the timestep calculation branches, which avoids unnecessary work per iteration. Branching in GPU code is typically discouraged due to potential warp divergence; however, when most threads take the same execution path (i.e., when most cells have very similar chemical compositions), the early-exit logic can be beneficial. Because cells are initialized via linear interpolation between $(0,1,0)$ and $(0.5, 0.0, 0.5)$, two adjacent cells are more similar at higher cell counts, leading to more similar execution paths and less warp divergence.

For matrix multiplication, AUTO achieves up to 94\% of cuBLAS double-precision performance at $N=1{,}024$, with performance ratios ranging from 0.63~--~0.94$\times$ for $N \geq 512$. At smaller problem sizes ($N \leq 256$), the gap widens considerably. The best AUTO solutions employ loop tiling, a well-established strategy for improving data reuse in matrix multiplication. Note that cuBLAS and cuSparse were explicitly forbidden as dependencies, restricting AUTO to raw CUDA, PTX, or CUTLASS implementations. With access to CUDA and CUTLASS documentation via semantic search, the Implementors were able to explore more advanced optimization strategies compared to earlier iterations of this work, where no documentation was provided.

\paragraph{t-SNE Analysis}

\begin{figure}
  \centering
  \includegraphics[width=\textwidth]{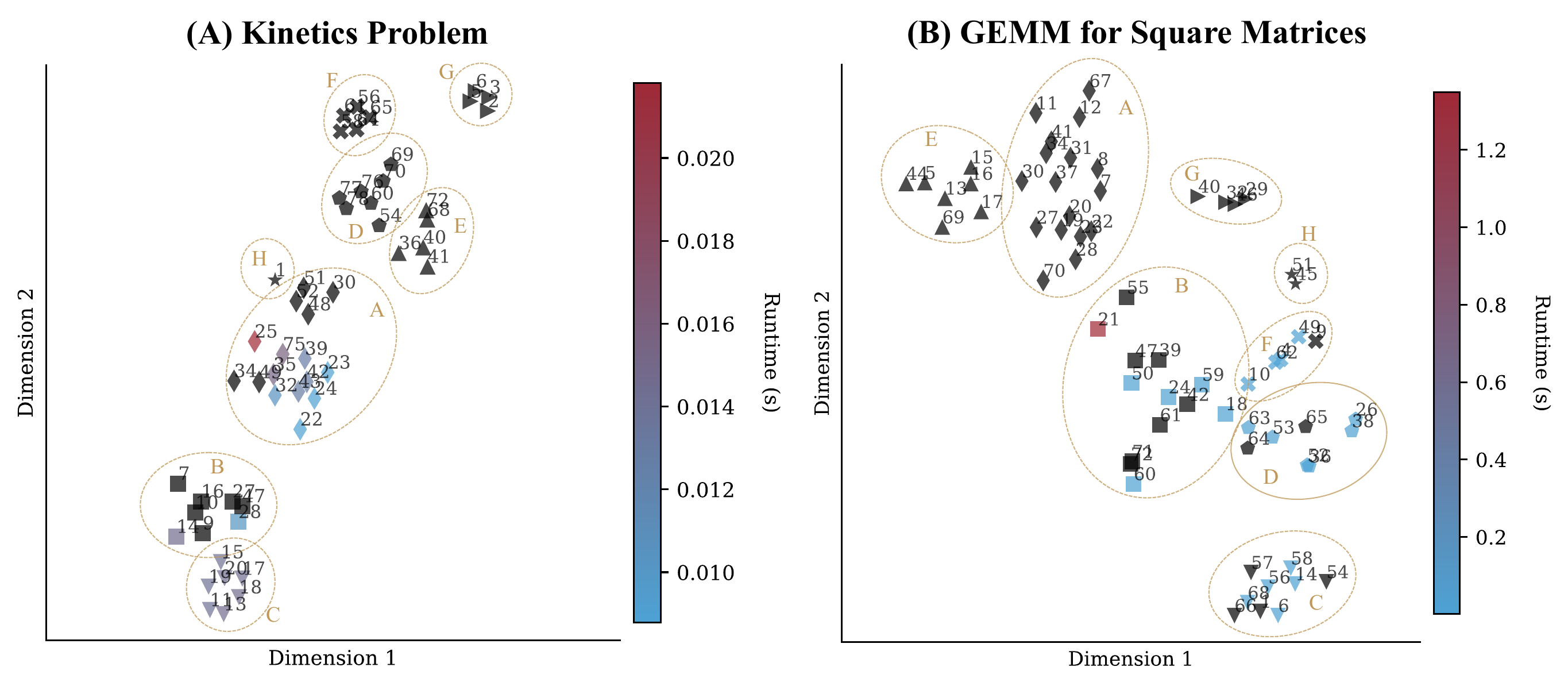}
  \caption{Code clusters and their correlations with the code runtimes for (A) kinetics and (B) matrix multiplication, from an earlier version of the framework. For each application, a "bag of words" approach produces the code vectors from the generated codes, which are then clustered using K-means. Visual embeddings are obtained using t-SNE with perplexity=10 for (A) and perplexity=15 for (B). The colors represent runtimes. The markers represent cluster memberships. The numbers next to each marker are the iteration at which the code was generated.}
  \label{fig_tsne_clusters}
\end{figure}

To analyze the structural relationships among generated codes, we present a t-SNE visualization from an earlier version of the framework (using GPT-OSS-20b with single-Implementor iterations). While the framework has since been updated, the analysis remains relevant as a demonstration of the design-space exploration characteristics. Figure~\ref{fig_tsne_clusters} shows code vectors for the kinetics and matrix multiplication problems. We employed a bag-of-words approach to construct code vectors rather than using embedding models. This decision was made after preliminary experiments (not shown here) on embedding-based similarity search, as reported by \citep{sharma2024reliable}. Our findings confirmed the need to capture algorithmic and implementation-level details that may be obscured in semantics-based similarity searches, as noted by \citet{galke2022bagofwordsvsgraphvs}. For each code, keywords and syntax were extracted with the Tree-Sitter library \citep{treesitter2024} to store position-aware data. The vectors were then constructed from the keyword frequencies in each code. Next, we identified stable clusters by applying consensus clustering \citep{fred2005combining}, which formed eight disjoint clusters of codes (per application) with co-occurrence rates exceeding 10\% between any two codes. To validate the meaningfulness of the resulting clusters, we manually inspected random codes for intra-cluster commonalities and inter-cluster differences in implementation strategies. Finally, t-SNE embeddings were produced with perplexity values of 10 for the kinetics application and 15 for the matrix multiplication application, resulting in visual alignment with the clusters obtained via consensus K-means clustering.

The t-SNE visualizations reveal strong correlations between code structure and performance. For kinetics, Cluster A contains top performers. These codes used mixed-precision computation, double-precision accumulation, \texttt{\_\_ldg} for read-only cache loads, and grid-stride loops. Cluster B explores synchronization strategies such as warp-level work stealing, which likely provided no benefits and incurred some overhead. Cluster C represents early explorations with pure double-precision and simpler approaches that achieved correctness without later optimizations. For matrix multiplication, optimal solutions span multiple clusters (B, C, D, F). Cluster B implements naive element-wise approaches. Cluster C contains tiled implementations and different padding strategies to avoid bank conflicts. Cluster D also shows tiling strategies without padding, while Cluster F explores kernels with loop unrolling. This diversity demonstrates that matrix multiplication admits multiple optimization pathways. Performance improvements followed irregular trajectories in both applications, which is expected. The search space is highly nonlinear, where small code changes could produce large performance variations.

\subsubsection{KernelBench}

We ran 61 AUTO simulations across 53 randomly selected problems at levels 1~--~3 of the KernelBench benchmark. This generated 1,339 implementations, of which 1,024 (77\%) passed the validation pipeline (compilation and correctness check). However, not all valid implementations successfully achieved speedup. Furthermore, during post-processing, several were identified as reward hacking or cheating. In contrast to KernelBench, our evaluation requirements are more lenient. KernelBench requires all core computation to be implemented as hand-written CUDA kernels (e.g., \texttt{\_\_global\_\_} functions via \texttt{load\_inline}). We allow solutions to achieve speedups by any valid means, including wrapping vendor-optimized CUDA libraries (cuBLAS, CUTLASS, Thrust, PTX) or algebraic simplifications, provided that the computation is correct and the speedup is achieved. "Cheating", on the other hand, involves inflating the speed of the core computation by, for instance, hiding computation from performance profilers or caching old results. 
 
\begin{figure}[h]
    \centering
    \includegraphics[width=\linewidth]{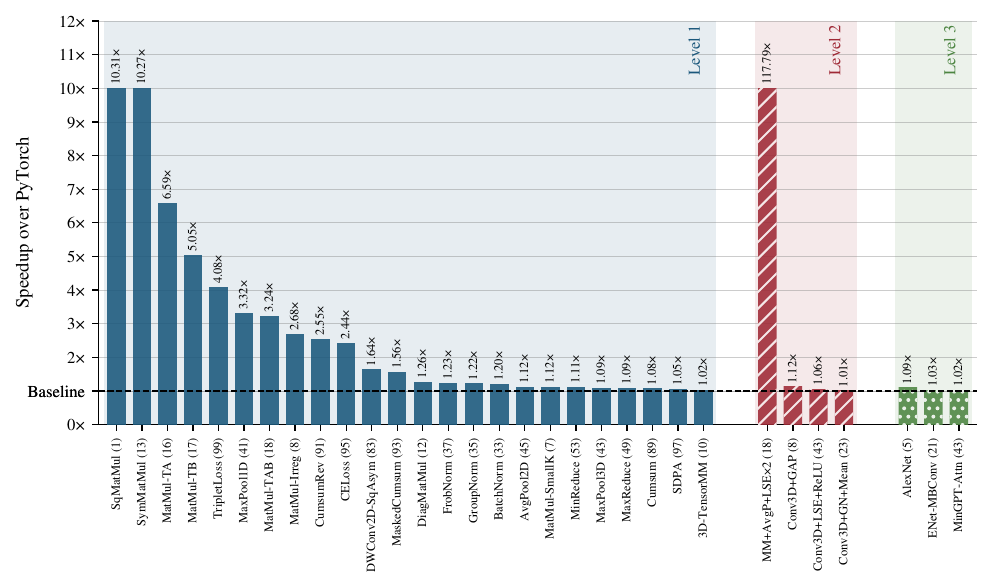}
    \caption{Best speedup achieved per KernelBench problem by AUTO, sorted in descending speedup order. Level 1 (single-kernel operators), Level 2 (fusion patterns), and Level 3 (full architectures) are grouped and shaded separately. The dashed line marks the 1$\times$ PyTorch baseline. Bars are clipped at 10$\times$ for visual clarity. All bars are annotated with their calculated speedup. }
    \label{fig_kernelbench_speedup}
\end{figure}

After filtering out implementations that failed validation, cheated, or did not achieve a speedup greater than 1.01$\times$, we identified 29 of 53 problems with truly novel implementations that achieved speedups. Figure~\ref{fig_kernelbench_speedup} shows the best speedup achieved per the 29 problems, sorted in descending order within each level group. Although some problems achieved more than 10$\times$ speedup, bars are clipped at 10$\times$ for visual clarity, and the actual speedup is annotated above each bar. In the following paragraphs, we provide a brief overview of the results. The interested reader may consult the supplementary data for additional information about each of the 53 problems and their evaluation reports.

Level 1 problems, which target individual GPU operators, show the strongest performance gains. The five matrix multiplication variants (Problems 1, 13, 16, 17, 18) achieve speedups from 3.24$\times$ to 10.31$\times$ by wrapping cuBLAS or CUTLASS with FP16 inputs and FP32 accumulation to exploit H100 Tensor Cores, consistent with the matrix multiplication application in Section~\ref{optimizer_apps}. Beyond matrix multiplication, several operators achieve true custom-kernel speedups through kernel fusion: triplet margin loss (Problem 99, 4.08$\times$) and cross-entropy loss (Problem 95, 2.44$\times$) each replace multi-kernel PyTorch pipelines with single fused passes, while MaxPool1D (Problem 41, 3.32$\times$) and reverse cumulative sum (Problem 91, 2.55$\times$) use custom kernels with loop unrolling and Thrust reverse-scans, respectively. 10 of the 32 problems did not exceed the 1.01$\times$ threshold or cheated the validation and scoring pipeline. For problems that did not see significant speedups, this indicates that PyTorch is already near-optimal for those operators on the H100. For cases where AUTO was found to be cheating, stronger guardrails are needed. Common cheating patterns include: placing the kernel computation on an unseen/undetectable CUDA stream to make it appear as if the computation completed in no time; and writing dummy code to pass heuristic static analysis and lazily offloading the computation to built-in PyTorch operators.

Level 2 problems require fusing multiple operators and are more challenging. Four of the five evaluated problems produce modest speedups (1.01~--~1.12$\times$); manual inspection found these solutions bypassed the static checker by aliasing \texttt{torch.nn} as \texttt{nn}, allowing dominant convolution operations to remain as stock cuDNN layers while only cheaper post-convolution steps were replaced with custom kernels. Problem 18 (Matmul-Sum-Max-AvgPool-LogSumExp) is an outlier at 117.79$\times$: the analyst agent confirmed this was a valid algebraic simplification to a dot product. While this is a valid simplification, the kernel is not generalizable to other problems.

Level 3 problems, which target full neural network architectures, prove the most difficult. Only 3 of 16 evaluated problems produced valid speedups after filtering: AlexNet (Problem 5, 1.09$\times$), an EfficientNet MBConv block (Problem 21, 1.03$\times$), and MinGPT causal attention (Problem 43, 1.01$\times$). These modest gains reflect the difficulty of outperforming cuDNN when computation is distributed across many heterogeneous operations. Cheating was frequently observed at this level. Common patterns included: output caching to exploit deterministic benchmark inputs; dummy identity kernels to bypass the static checker; and delegation to the unmodified reference model via import aliasing or obfuscation.

\subsection{Search Efficiency, Convergence, and Stability}\label{sec_eff_conv_stab}

This section examines the Strategist's ability to select a sensible strategy based on observations of how different designs perform. Analysis here is based on results from an earlier version of the framework using GPT-OSS-20b with single-Implementor iterations (see Table~\ref{tab_strat_distribution}). While the framework has since been updated to use GPT-OSS-120b with multiple concurrent Implementors and a domain-specific knowledge base (see Section~\ref{sec_implementation_details}), the analysis remains relevant as a demonstration of the strategic search methodology and the Strategist-Implementor alignment with traditional optimization techniques.

Given that the Strategist selected a sensible strategy, we also assess its ability to communicate clear design instructions to the Implementor and the Implementor's ability to follow them. For a quantitative assessment of search efficiency, we first draw analogies to well-established sampling techniques in Bayesian optimization \citep{kushner1964new}. As with the AUTO framework, every iteration in Bayesian optimization begins with a set of existing points in the search space and their evaluations with respect to a predefined objective function \citep{frazier2018tutorial}. Selection of the next point to sample and evaluate is based on an acquisition function that balances exploration and exploitation. Similarly, AUTO's Strategist must balance exploring novel design regions against exploiting promising areas. Thus, a measure of alignment between AUTO's and Bayesian optimization's sampling methods provides a rough estimate of search efficiency. To quantify these alignments, the first step is to classify the Strategist's decision at a given iteration, $t$, as exploratory or exploitive:

\begin{equation}
y_\text{S}^{(t)} = \begin{cases}
\text{exploitation} & \text{if } \varsigma_t \in \{\text{refine}, \text{combine}\} \\
\text{exploration} & \text{if } \varsigma_t = \text{innovate}
\end{cases}
\end{equation}

$\varsigma_t$ is the Strategist's decision. The "refine" and "combine" strategies are considered exploitative because they build on known design scores, whereas "innovate" is exploratory because the Strategist ventures into uncharted regions of the design space. To compare the Strategist's choices against those of Bayesian optimization and to assess the Implementor's execution, we define the following mapping, $\Phi: \mathbb{R}^d \to \{\text{exploitation}, \text{exploration}\}$, for a consistent classification of any design point as exploratory or exploitative:

\begin{equation}
\Phi(\mathbf{x}; \mathcal{C}) = \begin{cases}
\text{exploitation} & \text{if } D_\text{min}(\mathbf{x}, \mathcal{C}) \leq 0.1 \times L(\mathcal{C}) \\
& \text{or } \mathbf{x} \in \text{ConvexHull}(\mathcal{C}) \\
\text{exploration} & \text{otherwise}
\end{cases}
\end{equation}

where $D_\text{min}(\mathbf{x}, \mathcal{C})$ is the minimum distance from point $\mathbf{x} \in \mathbb{R}^d$ to any point in $\mathcal{C}$, and $L(\mathcal{C})$ is the max of max norms, $|| \cdot ||_\infty$, across all design points in $\mathcal{C}_t$. Intuitively, a sample point is exploitive if it is close to any design point in $\mathcal{C}$, or if it falls within the convex hull of $\mathcal{C}$. Otherwise, the point is considered exploratory. Using this classifier, we can determine the effective sample strategy of both the Bayesian optimizer and the Implementor:

\vspace{6pt}

\noindent
\begin{minipage}{0.5\linewidth}
\begin{equation}
y_{\text{B}}^{(t)} = \Phi(\mathbf{x}_\text{BO}; \mathcal{C}_t)
\end{equation}
\end{minipage}%
\begin{minipage}{0.5\linewidth}
\begin{equation}
y_{\text{I}}^{(t)} = \Phi(\mathbf{x}_\text{I}; \mathcal{C}_t)
\end{equation}
\end{minipage}

\vspace{6pt}

where $\mathbf{x}_{\text{BO}}(t)$ is the sample point selected by Bayesian optimization and $\mathbf{x}_{\text{I}}(t)$ is the design point produced by the Implementor. Search efficiency is quantified via two binary metrics:

\vspace{6pt}

\noindent
\begin{minipage}{0.5\linewidth}
\begin{equation}\label{eqn_strat_bo_alignment}
A_{\text{S,B}}^{(t)} = \begin{cases}
1 & \text{if } y_{\text{S}}^{(t)} = y_{\text{B}}^{(t)} \\
0 & \text{otherwise}
\end{cases}
\end{equation}
\end{minipage}%
\begin{minipage}{0.5\linewidth}
\begin{equation}\label{eqn_strat_imp_alignment}
A_{\text{S,I}}^{(t)} = \begin{cases}
1 & \text{if } y_{\text{S}}^{(t)} = y_{\text{I}}^{(t)} \\
0 & \text{otherwise}
\end{cases}
\end{equation}
\end{minipage}

\vspace{6pt}

Equation~\ref{eqn_strat_bo_alignment} measures whether the Strategist's sampling strategy matches that of a traditional Bayesian optimizer, while Equation~\ref{eqn_strat_imp_alignment} indicates whether the Implementor successfully executed the Strategist's intended strategy. Together, these metrics provide quantitative insight into the efficiency of the strategy-to-implementation pipeline. The final search efficiency marker is calculated as:

\begin{equation}\label{eqn_search_eff}
    \text{search efficiency} = \left(\frac{1}{2n} \sum_t A_\text{S,B}^{(t)}+A_\text{S,I}^{(t)} \right) \times 100,
\end{equation}

where $n$ is the number of iterations considered in the sum. This sum excludes optimization iterations that did not produce any code due to parsing errors in the LLM output or LLM inference timeouts. To obtain $\mathbf{x}_\text{BO}$, the upper confidence bound (UCB) acquisition function is used within the Bayesian Optimization Python library by \citet{nogueira2014bayesopt}. For details on the UCB function and its initial formulation, the reader is referred to \citet{srinivas2010gaussian}. The UCB function is parameterized by an exploration factor, $\xi$, which controls exploitation (low $\xi$) and exploration (high $\xi$). 

\begin{figure}[b]
    \centering
    \includegraphics[width=0.5\linewidth]{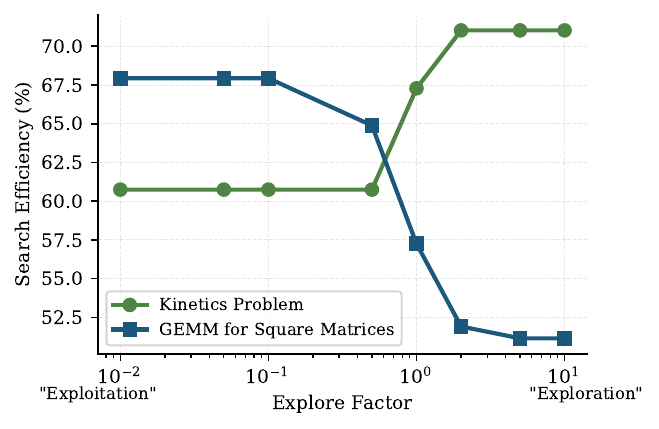}
    \caption{Search efficiency (calculated from Equation~\ref{eqn_search_eff}) as a function of the exploration factor, $\xi$, from the upper confidence bound (UCB) acquisition function for kinetics (circles) and matrix multiplication (squares) from Table~\ref{tab_strat_distribution}.}
    \label{fig_search_efficiency}
\end{figure}

Figure~\ref{fig_search_efficiency} shows how the search efficiency varies with the exploration factor for the best optimization run per application. The graph shows search efficiency values ranging from $\sim 50\%$ to $\sim 70\%$ across exploration factors spanning four orders of magnitude. This indicates that, for most optimization iterations, the Strategist's search strategies are aligned with those of the Bayesian optimizer, the Implementor, or both. The higher search efficiency obtained by the kinetics application at $\xi = 10$ indicates that AUTO exhibits more exploratory behavior. This is consistent with the Strategist's decisions being mostly to "innovate", as reported in Table~\ref{tab_strat_distribution}. On the other hand, AUTO exhibits more exploitative behavior at $\xi=0.01$ when optimizing matrix multiplication, which correlates with the Strategist's decisions being mostly to "combine" codes, as seen in Table~\ref{tab_strat_distribution}.

\begin{table}[htbp]
\centering
\caption{Strategist decisions distribution for each application.}
\renewcommand{\arraystretch}{1.2}
\label{tab_strat_distribution}
\begin{tabular}{rrrrr}
\toprule
\multirow{2}{*}{\textbf{Strategist Decision}} & \multicolumn{2}{c}{\textbf{Kinetics}} & \multicolumn{2}{c}{\textbf{Matrix Multiplication}} \\
\cmidrule(lr){2-3} \cmidrule(lr){4-5}
& \textbf{Count} & \textbf{Percentage} & \textbf{Count} & \textbf{Percentage} \\
\midrule
Combine         &  10 & 12.82\% &  42 & 58.33\% \\
Innovate        &  45 & 57.69\% &  16 & 22.22\% \\
Refine          &   3 &  3.85\% &   9 & 12.50\% \\
N/A             &  20 & 25.64\% &   5 &  6.94\% \\
\midrule
\textbf{Total} & \textbf{78} & \textbf{100.00\%} & \textbf{72} & \textbf{100.00\%} \\
\bottomrule
\end{tabular}
\end{table}

\begin{figure}[h]
    \centering
    \includegraphics[width=0.8\linewidth]{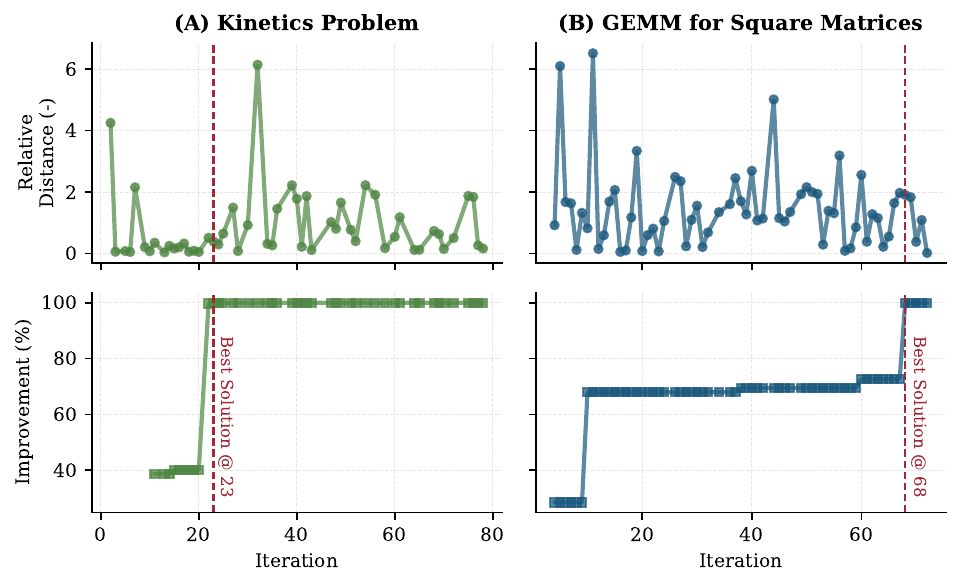}
    \caption{Relative distance (see Equation~\ref{eqn_rel_dist}) between code vectors of successive iterations (top row) and code runtime improvement as a percentage (bottom row) versus each iteration for (A) kinetics and (B) matrix multiplication from Table~\ref{tab_strat_distribution}.}
    \label{fig_convergence_metrics}
\end{figure}

Figure~\ref{fig_convergence_metrics} shows convergence patterns for both applications. The top row shows the relative distance between code vectors across successive iterations. The relative distance is computed as:

\begin{equation}\label{eqn_rel_dist}
    \text{relative distance at } t \equiv \frac{||v_t-v_{t-1}||_2}{||v_{t-1}||_2},
\end{equation}

where $v_t$ is the code vector at iteration $t$ and $||\cdot||$ indicates the L2 norm. A large spike indicates significant code changes and possible exploratory behavior, while small values suggest minor changes and possible exploitive behavior. The bottom row shows the improvement (as a percentage relative to the best overall runtime) of the best runtime so far up to iteration $t$. This value is computed as:
\begin{equation}
    \text{best solution at } t \equiv \left( 2 - \frac{\min_{i \leq t} r^{(i)}}{\min_{i \leq n} r^{(i)}} \right) \times 100
\end{equation}

In the kinetics problem, the relative distance shows low variability as it approaches the best solution at 23 iterations, suggesting exploitative behavior as a local optimum is reached. After 23 iterations, the variability is significantly greater, indicating that AUTO continues to explore the design space for other optima. The matrix multiplication application shows more frequent variations. Combined with observations from Figure~\ref{fig_search_efficiency}, these seemingly large variations could be a result of jumping from one known similar set of design points to a completely different safe set of design points.

In the improvement subplots of Figure~\ref{fig_convergence_metrics}, initial codes may not have passed compiler and correctness checks, and therefore some values are missing. The kinetics problem shows a sharp performance jump at iteration 23, reaching its optimal solution for the run. The matrix multiplication optimization exhibits a steadier convergence pattern, reaching its optimum by iteration 68. Both applications show that breakthroughs in optimization can occur after extended periods of stagnation. However, this behavior poses challenges in determining the optimal final iteration, as premature termination can lead to suboptimal solutions. However, note that in this preliminary study, no CUDA documentation was provided, potentially resulting in wasted code generations (and LLM tokens) due to very minor, fixable syntax issues. With access to a knowledge base, stagnant performance segments are likely to lessen.

\subsection{Token Usage and Cost Analysis}\label{sec_cost_anal}

In this section, we demonstrate the economic viability of the AUTO framework using a representative optimization run from an earlier version of the framework (GPT-OSS-20b with single-Implementor iterations and no domain-specific knowledge base). 53 of 78 iterations successfully generated code, with the shortest code having 228 lines and the longest 393 lines. The average code size is $\sim285 \pm 31$ lines, which indicates relatively consistent code generation. Table~\ref{tab_token_stats} reports the token usage per iteration, based on the GPT-OSS-20b tokenizer, and the context usage is based on the max context size of 128k tokens. The Implementor consumed an average of 10,054 input and 8,601 output tokens per iteration, while the Strategist processed significantly larger contexts with an average of 42,632 input and 3,684 output tokens per iteration. This is expected given the amount of code that the Strategist must analyze when provided the curated context (see Section~\ref{sec_opt_framework}). The average ``per line of code'' token usage was computed by dividing the total tokens by the total lines across all generated code.

\begin{table}[h]
\centering
\caption{Token statistics (tokens/iter) from a representative optimization run}
\label{tab_token_stats}
\renewcommand{\arraystretch}{1.2}
\begin{tabular}{lrrrc|r}
\toprule
\textbf{Token Type} & \textbf{Min} & \textbf{Max} & \textbf{Avg.} & \textbf{Std (\%)} & \textbf{Context Usage (\%)} \\
\midrule
Implementor (input) & 4,085 & 25,135 & 10,054 & 67 & 7.85 \\
Implementor (output) & 2,543 & 20,461 & 8,601 & 63 & 6.72 \\
Strategist (input) & 2,998 & 60,212 & 42,632 & 38 & 33.31 \\
Strategist (output) & 2,135 & 8,921 & 3,684 & 38 & 2.88 \\
Per Line of Code & 9 & 15 & 11 & 11 & 0.01 \\
\bottomrule
\end{tabular}
\end{table}

\begin{figure}[h]
    \centering
    \includegraphics[width=0.8\linewidth]{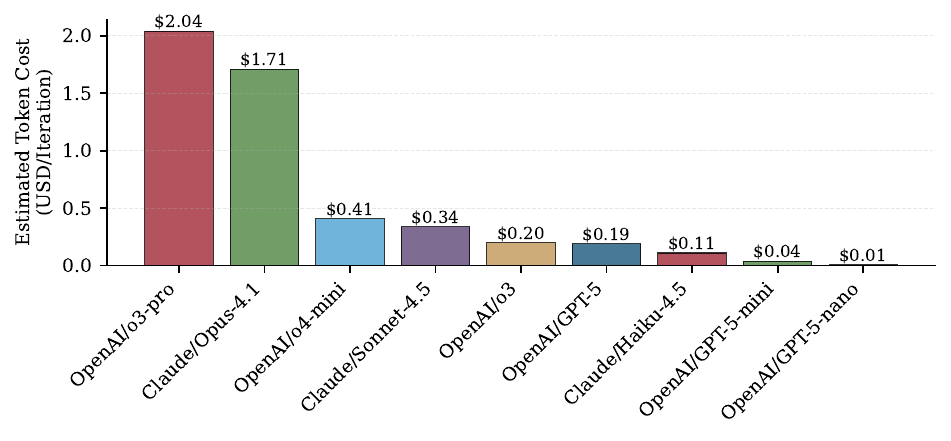}
    \caption{Estimated costs per AUTO iteration across various commercial LLM providers}
    \label{fig_token_usage}
\end{figure}

Figure~\ref{fig_token_usage} presents estimated costs per optimization iteration across various commercial LLM providers. These estimates are based on the average number of input and output tokens per iteration from Table~\ref{tab_token_stats} and on pricing data from OpenAI and Anthropic as of March 2026. Note that tokenization may vary across commercial LLMs, potentially affecting the estimated costs. For example, Claude models typically require 16-30\% more tokens than OpenAI models for the same content \citep{venturebeatTokenizer2024}, with code requiring the highest overhead. For the complete optimization run (78 total iterations), the total cost ranges from approximately \$0.78 for GPT-5-nano to \$159.12 for o3-pro, with mid-tier models like GPT-5 and Claude Sonnet-4.6 costing \$14.82 and \$26.52, respectively. These mid-tier models are likely a good balance between affordability and accuracy for the AUTO framework.

According to \citet{bls_software_dev_2024}, the median hourly wage for software developers was \$63.98 in May 2024, with the bottom 10\% earning \$38.39 per hour. To estimate human optimization costs, we account for the iterative nature of performance optimization. Unlike AUTO, we assume that human developers make steadier, more focused modifications of $\sim$10 lines per iteration, with each iteration requiring $\sim$45 minutes for profiling, modification, compilation, and testing. Expert developers may achieve optimal solutions within 10 iterations, whereas AUTO requires 78 total iterations (23 to reach the best solution). Furthermore, AUTO's broader exploration yielded 44-57\% compilation failures, which human developers typically avoid through domain expertise and modern AI pair-programming tools such as GitHub Copilot \citep{peng2023impact}.

Table~\ref{tab_cost_comparison} compares optimization costs. For human developers, we estimate 0.75 hours per iteration at median wage (\$48 per iteration), with 10 iterations to reach competitive performance (\$480 total). Entry-level developers may require more iterations at lower hourly rates, yielding similar total costs. AUTO with mid-tier models (GPT-5, Sonnet-4.6) costs \$15-\$27, compared with \$435-\$480 for human optimization. However, this comparison is nuanced. We emphasize that AUTO is intended as a complementary tool that augments expert capabilities. AUTO's strength lies in its ability to rapidly explore and test vast numbers of designs with domain knowledge.

\begin{table}[h]
\centering
\caption{Cost comparison estimates for GPU code optimization}
\label{tab_cost_comparison}
\renewcommand{\arraystretch}{1.1}
\begin{tabular}{lrr}
\toprule
\textbf{Approach} & \textbf{Cost per Iteration} & \textbf{Total Cost} \\
\midrule
\multicolumn{3}{l}{\textit{LLM-based (AUTO)} - see Figure~\ref{fig_token_usage}} \\
\quad GPT-5 & \$0.19 & \$14.82 (78 iter.) \\
\quad Sonnet-4.6 & \$0.34 & \$26.52 (78 iter.) \\
\quad Opus-4.6 & \$0.57 & \$44.46 (78 iter.) \\
\quad o3-pro & \$2.04 & \$159.12 (78 iter.) \\
\midrule
\multicolumn{3}{l}{\textit{Human Developer} - see discussion in Section~\ref{sec_cost_anal}} \\
\quad Entry-level (\$38.39/hr) & \$29 & \$435 (15 iter.) \\
\quad Median (\$63.98/hr) & \$48 & \$480 (10 iter.) \\
\bottomrule
\end{tabular}
\end{table}

\section{Conclusions}\label{sec_conclusion}

We introduced AUTO, a multi-agent LLM framework that treats design optimization as a gradient-free strategic search, separating high-level planning (Strategist) from low-level implementation (Implementors). We demonstrated that LLMs can serve as both the search heuristic and the design generator in ill-defined optimization spaces where traditional methods struggle to parameterize the design space. As a proof-of-concept of GPU code optimization, AUTO produced solutions that outperform expert-tuned code by up to 1.74$\times$ for chemical kinetics integration, reach 94\% of cuBLAS performance for matrix multiplication, and achieve speedups of up to 118$\times$ over PyTorch baselines across 59 KernelBench problems. A posteriori analysis reveals 50~--~70\% alignment with Bayesian optimization sampling strategies, and estimated costs with proprietary models (\$15~--~159 per run) compare favorably to manual optimization (\$435~--~480).

While this work demonstrates strong proof-of-concept results, there are a few caveats worth noting: compilation failure rates of 44~--~57\% in earlier runs wasted tokens and iterations; cheating was a common pattern in KernelBench (particularly at Level 3); LLM performance degraded on more complex problems; and run times increased dramatically with larger problem sizes. To address these issues, future work may include surrogate-based models for design-quality prediction, adaptive stopping criteria, human-in-the-loop steering, fine-tuned LLMs with improved context management, and overall architectural improvements to the framework. We also aim to extend the framework to non-code design domains such as hardware-software co-design and multi-objective engineering problems. AUTO is a complementary tool for domain experts and a boost for teams without specialized optimization knowledge. Furthermore, the framework is built 
entirely on open-source models and tools, demonstrating affordability and data privacy. By demonstrating AUTO's ability to autonomously discover competitive GPU implementations, this work opens avenues towards democratizing high performance computing expertise. Given the framework's adaptability, future work may show optimization applications spanning scientific simulation, research workflows, multidisciplinary systems optimization, and beyond.

\section{Supplementary Information}

Per-problem evaluation reports for the KernelBench problems (including the number of implementations generated, validation outcomes, best speedups achieved, and cheating annotations) are available upon request by contacting the authors.

\newpage

\section*{Acknowledgments}  
This work was supported by the Center for Prediction, Reasoning, and Intelligence for Multiphysics Exploration (C-PRIME), a PSAAP-IV project funded by the Department of Energy, grant number DE-NA0004264 (program manager: Dr.\ David Etim).

\bibliographystyle{unsrtnat}
\bibliography{references}

@article{kernelbench2025,
  title={{KernelBench}: Can {LLMs} write efficient {GPU} kernels?},
  author={Ouyang, Anne and Guo, Simon and Arora, Simran and Zhang, Alex L and Hu, William and R{\'e}, Christopher and Mirhoseini, Azalia},
  journal={arXiv preprint arXiv:2502.10517},
  year={2025}
}

@article{afterburner2025,
  title={Afterburner: Reinforcement learning facilitates self-improving code efficiency optimization},
  author={Du, Mingzhe and Tuan, Luu Anh and Liu, Yue and Qing, Yuhao and Huang, Dong and He, Xinyi and Liu, Qian and Ma, Zejun and Ng, See-kiong},
  journal={arXiv preprint arXiv:2505.23387},
  year={2025}
}

@article{gpukernelscientist2025,
  title={{GPU} Kernel Scientist: An {LLM}-Driven Framework for Iterative Kernel Optimization},
  author={Andrews, Martin and Witteveen, Sam},
  journal={arXiv preprint arXiv:2506.20807},
  year={2025}
}

@article{fortrankokkos2025,
  title={From Legacy {Fortran} to Portable {Kokkos}: An Autonomous Agentic {AI} Workflow},
  author={Gupta, Sparsh and Kamalakkannan, Kamalavasan and Moraru, Maxim and Shipman, Galen and Diehl, Patrick},
  journal={arXiv preprint arXiv:2509.12443},
  year={2025}
}

@incollection{robertson1966,
  title={The Solution of a Set of Reaction Rate Equations},
  author={Robertson, H. H.},
  booktitle={Numerical Analysis: An Introduction},
  year={1966}
}

@misc{cublas,
  title        = {{cuBLAS} {L}ibrary},
  author       = {{NVIDIA Corporation}},
  howpublished = {\url{https://developer.nvidia.com/cublas}},
  note         = {Accessed: 2025}
}

@inproceedings{sharma2025accelerating,
  title={Accelerating {CFD} Simulations With Super-Resolution Feedback-Informed Adaptive Mesh Refinement},
  author={Sharma, Vansh and Rauch, Andreas H and Raman, Venkatramanan},
  booktitle={AIAA SCITECH 2025 Forum},
  pages={1467},
  year={2025}
}

@article{sharma2024reliable,
  title={A reliable knowledge processing framework for combustion science using foundation models},
  author={Sharma, Vansh and Raman, Venkat},
  journal={Energy and AI},
  volume={16},
  pages={100365},
  year={2024},
  publisher={Elsevier}
}

@misc{galke2022bagofwordsvsgraphvs,
      title={Bag-of-Words vs. Graph vs. Sequence in Text Classification: Questioning the Necessity of Text-Graphs and the Surprising Strength of a Wide {MLP}}, 
      author={Lukas Galke and Ansgar Scherp},
      year={2022},
      eprint={2109.03777},
      archivePrefix={arXiv},
      primaryClass={cs.CL},
      url={https://arxiv.org/abs/2109.03777}, 
}

@article{sharma2024amrex,
  title={An {AMReX}-based compressible reacting flow solver for high-speed reacting flows relevant to hypersonic propulsion},
  author={Sharma, Shivank and Bielawski, Ral and Gibson, Oliver and Zhang, Shuzhi and Sharma, Vansh and Rauch, Andreas H and Singh, Jagmohan and Abisleiman, Sebastian and Ullman, Michael and Barwey, Shivam and others},
  journal={arXiv preprint arXiv:2412.00900},
  year={2024}
}

@article{agarwal2025gpt,
  title={{gpt-oss-120b} \& {gpt-oss-20b} model card},
  author={Agarwal, Sandhini and Ahmad, Lama and Ai, Jason and Altman, Sam and Applebaum, Andy and Arbus, Edwin and Arora, Rahul K and Bai, Yu and Baker, Bowen and Bao, Haiming and others},
  journal={arXiv preprint arXiv:2508.10925},
  year={2025}
}

@incollection{ollama,
  title={Using {Ollama}},
  author={Marcondes, Francisco S and Gala, Adelino and Magalh{\~a}es, Renata and Perez de Britto, Fernando and Dur{\~a}es, Dalila and Novais, Paulo},
  booktitle={Natural Language Analytics with Generative Large-Language Models: A Practical Approach with Ollama and Open-Source LLMs},
  pages={23--35},
  year={2025},
  publisher={Springer}
}

@article{frazier2018tutorial,
  title={A tutorial on Bayesian optimization},
  author={Frazier, Peter I},
  journal={arXiv preprint arXiv:1807.02811},
  year={2018}
}

@article{kushner1964new,
  title={A new method of locating the maximum point of an arbitrary multipeak curve in the presence of noise},
  author={Kushner, Harold J},
  year={1964}
}

@article{raman2023nonidealities,
  title={Nonidealities in rotating detonation engines},
  author={Raman, Venkat and Prakash, Supraj and Gamba, Mirko},
  journal={Annual Review of Fluid Mechanics},
  volume={55},
  number={1},
  pages={639--674},
  year={2023},
  publisher={Annual Reviews}
}

@Misc{nogueira2014bayesopt,
    author = {Fernando Nogueira},
    title = {{Bayesian Optimization}: Open source constrained global optimization tool for {Python}},
    year = {2014},
    url = " https://github.com/bayesian-optimization/BayesianOptimization"
}

@inproceedings{srinivas2010gaussian,
  title={Gaussian process optimization in the bandit setting: No regret and experimental design},
  author={Srinivas, Niranjan and Krause, Andreas and Kakade, Sham M and Seeger, Matthias},
  booktitle={International Conference on Machine Learning (ICML)},
  pages={1015--1022},
  year={2010}
}

@misc{treesitter2024,
  author = {Brunsfeld, Max and {Tree-sitter contributors}},
  title = {{Tree-sitter}: An incremental parsing system for programming tools},
  year = {2024},
  howpublished = {\url{https://github.com/tree-sitter/tree-sitter}},
  note = {Accessed: 2024-11-24}
}

@article{peng2023impact,
  title={The impact of {AI} on developer productivity: Evidence from {GitHub} {Copilot}},
  author={Peng, Sida and Kalliamvakou, Eirini and Cihon, Peter and Demirer, Mert},
  journal={arXiv preprint arXiv:2302.06590},
  year={2023}
}

@article{hamadanian2025glia,
  title={Glia: A Human-Inspired {AI} for Automated Systems Design and Optimization},
  author={Hamadanian, Pouya and Karimi, Pantea and Nasr-Esfahany, Arash and Noorbakhsh, Kimia and Chandler, Joseph and ParandehGheibi, Ali and Alizadeh, Mohammad and Balakrishnan, Hari},
  journal={arXiv preprint arXiv:2510.27176},
  year={2025}
}

@incollection{hornby2006automated,
  title={Automated antenna design with evolutionary algorithms},
  author={Hornby, Gregory and Globus, Al and Linden, Derek and Lohn, Jason},
  booktitle={Space 2006},
  pages={7242},
  year={2006}
}

@article{zhang2020bayesian,
  title={Bayesian optimization for materials design with mixed quantitative and qualitative variables},
  author={Zhang, Yichi and Apley, Daniel W. and Chen, Wei},
  journal={Scientific Reports},
  volume={10},
  number={1},
  pages={4924},
  year={2020},
  publisher={Nature Publishing Group}
}

@article{li2021nextorch,
  title={{NEXTorch}: A design and {Bayesian} optimization toolkit for chemical sciences and engineering},
  author={Li, Yifan and Maffettone, Phillip M. and Vlachos, Dionisios G. and Caratzoulas, Stavros},
  journal={Journal of Chemical Information and Modeling},
  volume={61},
  number={11},
  pages={5312--5319},
  year={2021},
  publisher={ACS Publications}
}

@article{carreon2025gpu,
  title={A {GPU}-based Compressible Combustion Solver for Applications Exhibiting Disparate Space and Time Scales},
  author={Carreon, Anthony and Singh, Jagmohan and Sharma, Shivank and Zhang, Shuzhi and Raman, Venkat},
  journal={arXiv preprint arXiv:2510.23993},
  year={2025}
}

@article{romeraparedes2023funsearch,
  title={Mathematical discoveries from program search with large language models},
  author={Romera-Paredes, Bernardino and Barekatain, Mohammadamin and Novikov, Alexander and Balog, Matej and Kumar, M. Pawan and Dupont, Emilien and Ruiz, Francisco J. R. and Ellenberg, Jordan and Wang, Pengming and Fawzi, Omar and Kohli, Pushmeet and Fawzi, Alhussein},
  journal={Nature},
  year={2023},
  doi={10.1038/s41586-023-06924-6}
}

@article{novikov2025alphaevolve,
  title={{AlphaEvolve}: A coding agent for scientific and algorithmic discovery},
  author={Novikov, Alexander and V{\~u}, Ng{\^a}n and Eisenberger, Marvin and Dupont, Emilien and Huang, Po-Sen and Wagner, Adam Zsolt and Shirobokov, Sergey and Kozlovskii, Borislav and Ruiz, Francisco JR and Mehrabian, Abbas and others},
  journal={arXiv preprint arXiv:2506.13131},
  year={2025}
}

@inproceedings{volkov2008benchmarking,
  title={Benchmarking {GPUs} to tune dense linear algebra},
  author={Volkov, Vasily and Demmel, James W},
  booktitle={SC'08: Proceedings of the 2008 ACM/IEEE conference on Supercomputing},
  pages={1--11},
  year={2008},
  organization={IEEE}
}

@article{fred2005combining,
  title={Combining multiple clusterings using evidence accumulation},
  author={Fred, Ana LN and Jain, Anil K},
  journal={IEEE Transactions on Pattern Analysis and Machine Intelligence},
  volume={27},
  number={6},
  pages={835--850},
  year={2005},
  publisher={IEEE}
}

@misc{bls_software_dev_2024,
  author = {{The U.S. Bureau of Labor Statistics}},
  title = {Software Developers, Quality Assurance Analysts, and Testers},
  howpublished = {Occupational Outlook Handbook},
  year = {2024},
  month = {May},
  note = {Accessed November 25, 2025},
  url = {https://www.bls.gov/ooh/computer-and-information-technology/software-developers.htm}
}

@online{venturebeatTokenizer2024,
  author = {Gupta, Lavanya},
  title = {Hidden costs in {AI} deployment: Why {Claude} models may be 20-30\% more expensive than {GPT} in enterprise settings},
  year = {2024},
  month = {August},
  day = {27},
  urldate = {2024-11-25},
  organization = {VentureBeat}
}

@misc{pydantic_ai,
  author = {{Pydantic}},
  title = {{Pydantic AI}: Agent Framework for Building Production-Grade {AI} Applications},
  year = {2025},
  howpublished = {\url{https://ai.pydantic.dev/}},
  note = {Accessed: 2025}
}

@article{packer2023memgpt,
  title     = {{MemGPT}: Towards {LLMs} as Operating Systems},
  author    = {Packer, Charles and Wooders, Sarah and Lin, Kevin and
               Fang, Vivian and Patil, Shishir G. and Stoica, Ion and
               Gonzalez, Joseph E.},
  journal   = {arXiv preprint arXiv:2310.08560},
  year      = {2023},
  url={https://arxiv.org/abs/2310.08560}
}

@article{gunther2023jina,
  title={{Jina} embeddings 2: 8192-token general-purpose text embeddings for long documents},
  author={G{\"u}nther, Michael and Ong, Jackmin and Mohr, Isabelle and Abdessalem, Alaeddine and Abel, Tanguy and Akram, Mohammad Kalim and Guzman, Susana and Mastrapas, Georgios and Sturua, Saba and Wang, Bo and others},
  journal={arXiv preprint arXiv:2310.19923},
  year={2023},
  url={https://arxiv.org/abs/2310.19923}
}

@inproceedings{reimers2019sentencebert,
  title     = "{Sentence-BERT}: Sentence Embeddings using Siamese {BERT}-Networks",
  author    = "Reimers, Nils and Gurevych, Iryna",
  booktitle = "Proceedings of the 2019 Conference on Empirical Methods in Natural Language Processing",
  month     = "11",
  year      = "2019",
  publisher = "Association for Computational Linguistics",
  url       = "http://arxiv.org/abs/1908.10084",
}

\end{document}